\documentclass[sigconf]{acmart}
\usepackage{hyperref}
\usepackage{cleveref}
\usepackage{booktabs} 
\usepackage{multirow}
\usepackage{color}
\usepackage{enumitem}

\setcopyright{rightsretained}

\acmDOI{10.475/123_4}

\acmISBN{123-4567-24-567/08/06}

\acmYear{2019}
\copyrightyear{2019}
\acmArticle{4}
\acmPrice{15.00}

\begin{document}

\fancyhead{}
\title{Time-weighted Attentional Session-Aware Recommender System}

\author{Mei Wang}
\affiliation{%
  \institution{University of Texas at Austin}
  \city{Austin, TX}
  \country{USA}
}
\email{meiwang@cs.utexas.edu}

\author{Weizhi Li}
\affiliation{%
  \institution{JD.COM.}
  \city{Mountain View, CA}
  \country{USA}
}
\email{weizhi.li@jd.com}

\author{Yan Yan}
\affiliation{%
  \institution{Facebook Inc}
  \city{Menlo Park, CA}
  \country{USA}}
\email{chrisyan@fb.com}
\renewcommand{\shorttitle}{}
\renewcommand{\shortauthors}{}

\begin{abstract}
Session-based Recurrent Neural Networks (RNNs) are gaining increasing popularity for recommendation task, due to the high autocorrelation of user's behavior on the latest session and the effectiveness of RNN to capture the sequence order information.
However, most existing session-based RNN recommender systems still solely focus on the short-term interactions within a single session and completely discard all the other long-term data across different sessions. While traditional Collaborative Filtering (CF) methods have many advanced research works on exploring long-term dependency, which show great value to be explored and exploited in deep learning models. 
Therefore, in this paper, we propose ASARS, a novel framework that effectively imports the temporal dynamics methodology in CF into session-based RNN system in DL, such that the temporal info can act as scalable weights by a parallel attentional network.
Specifically, we first conduct an extensive data analysis to show the distribution and importance of such temporal interactions data both within sessions and across sessions. 
And then, our ASARS framework promotes two novel models: 
(1) an inter-session temporal dynamic model that captures the long-term user interaction for RNN recommender system. We integrate the time changes in session RNN and add user preferences as model drifting;
and (2) a novel triangle parallel attention network that enhances the original RNN model by incorporating time information. Such triangle parallel network is also specially designed for realizing data argumentation in sequence-to-scalar RNN architecture, and thus it can be trained very efficiently.
Our extensive experiments on four real datasets from different domains demonstrate the effectiveness and large improvement of ASARS for personalized recommendation.
\end{abstract}

\settopmatter{printacmref=false}

\keywords{Recommender System, Session-based RNN, Time-weighted Attention, Short-term and Long-term Profile}

\maketitle

\section{Introduction}

Recommender Systems (RS) have long been developed to predict user's favorites, evolving from traditional Collaborative Filtering (CF) methods \cite{schafer2007collaborative, koren2009matrix, mnih2008probabilistic, rendle2012factorization} to recently grown popular Deep Learning approaches \cite{covington2016deep, cheng2016wide, sedhain2015autorec}. 
As online services like e-commerce (Amazon), social media (Facebook), movie (YouTube) and music (Spotify) grow in an increasing speed of rate, how to improve recommendation quality from such expanding and wide-ranging items is prominent for both user experience and business profit.

\begin{figure*}[t] 
\includegraphics[width=0.32\textwidth]{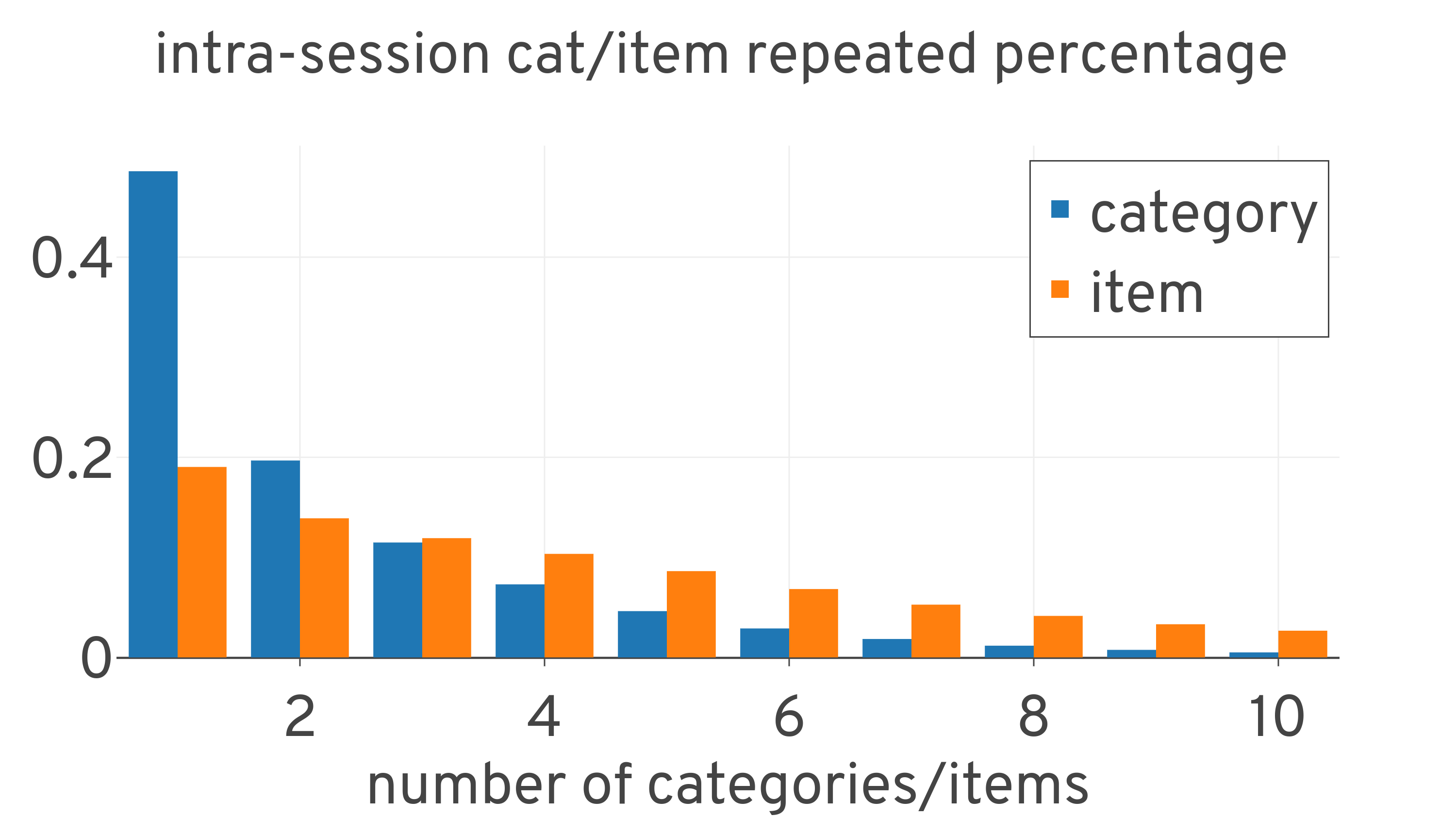}
\includegraphics[width=0.32\textwidth]{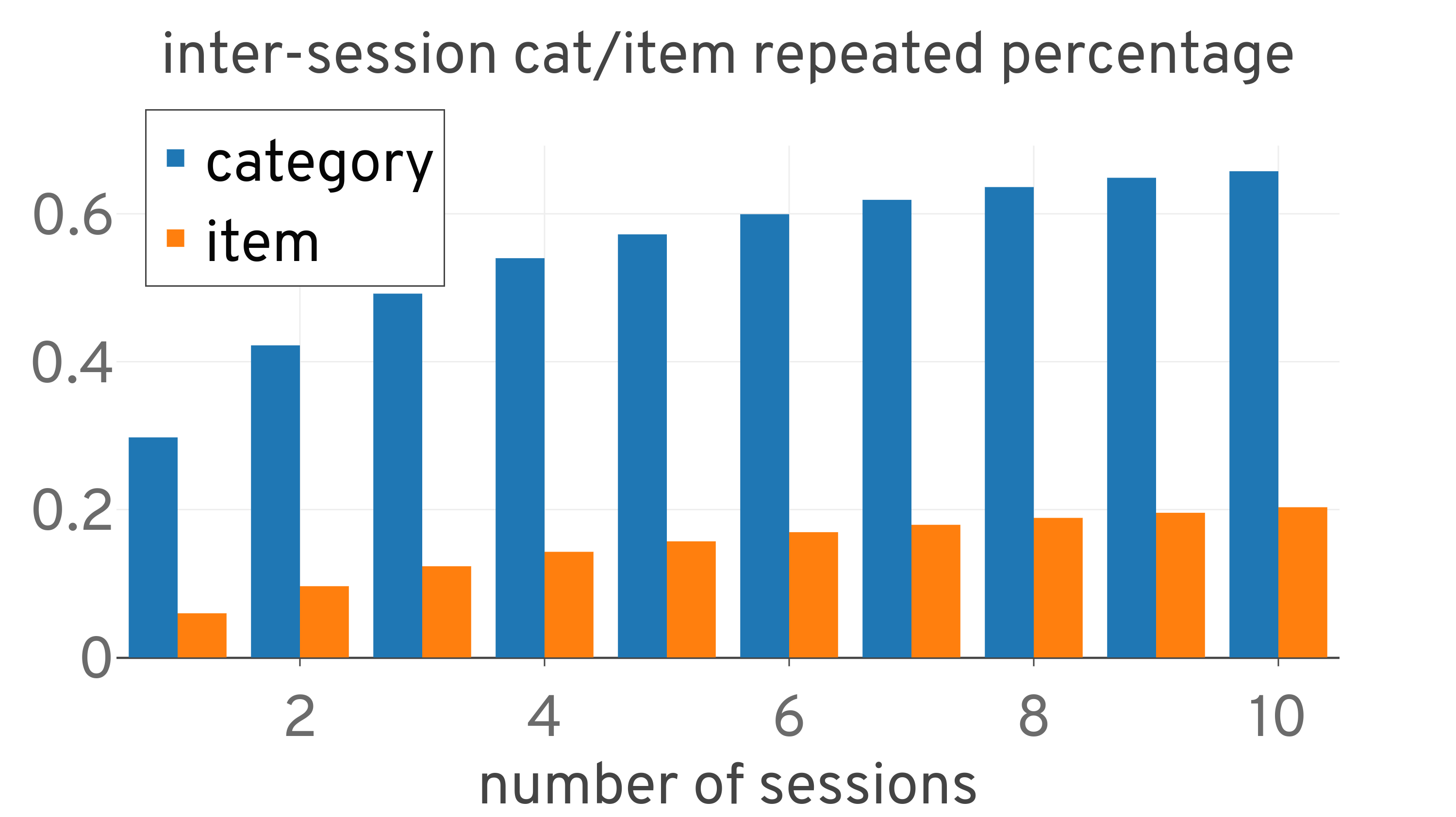}
\includegraphics[width=0.32\textwidth]{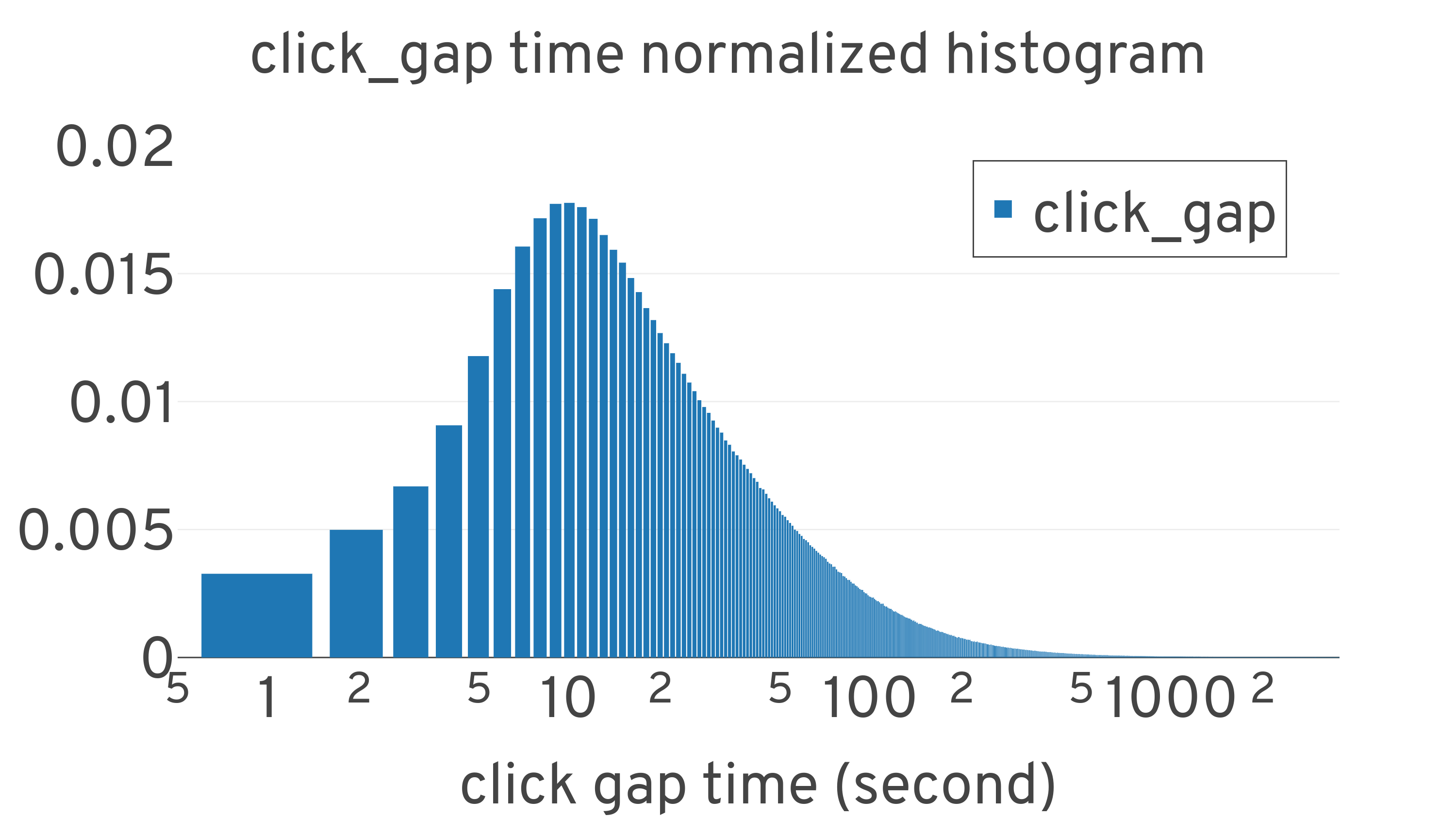}
\caption{An empirical data analysis: (1) Short-term predominates: the mean percentage of user interactions hanging in the top 10 categories/items during the same session;
(2) Long-term counts: the mean percentage value of user clicking the repeated categories and items that he had clicked before in the previous 10 sessions;
(3) Dwell Time helps: the normalized histogram of click gap follows gamma distribution.}
\label{fig:EDA}
\end{figure*}

\subsection{Personalized Recommender Systems}
Ever-more importantly, personalized recommendation is one of the most challenging issues in RS. 
User's intent is more than difficult to predict, which can be influenced by many factors, both internal or external, from past or current.
Much research efforts have focused on Context-Aware RS \cite{adomavicius2011context}, exploring contextual data like temporal information, spatial location \cite{Park2007LocationBasedRS, ge2016taper}, user profiles \cite{Adomavicius2005IncorporatingCI, rendle2009bpr} or even inter-domain features \cite{van2004context, reddy2006lifetrak}. 
Among them, Time-Aware RS \cite{campos2014time} is particularly studied in depth, since temporal information is easy-to-obtain and indicative of user's information need. 
Typically, temporal dynamics was added in CF methods to discover temporal evolving features \cite{koren2009collaborative} and many other sophisticated NN models were proposed, like time gates, point process, multi-task \cite{zhu2017next, dai2016recurrent, Shen2018MultiTaskLF} etc.
What's more, the ordering of interactions is another new dimension of information that is valuable to be further explored. 
Recent works \cite{hidasi2015session, tan2016improved, hidasi2017recurrent} show that RNN-based recommender system can outperform other popular alternatives in certain session-based recommendation tasks.

Sequential interactions between users and items are crucial data sources for recommender systems.
However, literatures above fail to quantify the effectiveness of using such sequential data from the past to present sessions.
Users' intents are constantly evolving and same as the item popularities. 
As a result, they cannot be effectively modeled based solely on short-term or long-term profiles. 

\noindent \textbf{Example} In an e-commerce recommender system, a user Alice may come with a certain intent for some kitchen hand soap that she wants to buy at present. Then in this current session, Alice is more likely to click some similar items or accessories from the same kitchenware category, like kitchen caddy, drying tower, trash bags, etc.
This means such short-term intra-session data sequence should mostly play a dominant role for the next-basket recommendation. 
At the same time, the dwell time she spent on each items could indicate the probability of her interest of making a purchase. 
On the other hand, her long-term profile of tastes or preferences would not change much over recent sessions, like her favorite brands, preferred color, fashion style, etc. 
Moreover, the items she viewed or bought in the past sessions may still give hints for the next session recommendation. For instance, fast moving consumer goods, makeups and napkins have periodic purchase needs. 
So maintaining both user's short-term intra-session context profile and long-term inter-session preference profile can lead to significant recommendation performance increase.

The goal of this work is to make effective use of both intra-session and inter-session profiles and to construct a better personalized session-aware recommender system.
This raises several challenging issues. 
First, traditional RNN cannot train with too long sequence length, which will result in extreme training latency and large memory cost. 
Second, the interaction data is very noisy: some clicks are meaningful, 
while some may even be clicked by mistake. 
Last but not least, data from the past sessions should play as different roles as present session, but there is no specific rule for integrating the session-based short-term profiles and session-aware long-term profiles.
Therefore, a more sensitive approach to distinguish and integrate data from different time scales with different significance is required.

\subsection{Motivated by Empirical Data Analysis}
\label{sec:EDA}
The motivation of our model design is inspired by real data observations and analysis, which comes from online Tianchi \cite{yi2015purchase} e-commerce navigation log data having around 100M interactions, 1M users and 4M items from 10K categories. We come up with the following three observations:
\begin{enumerate}[leftmargin=*]
\item{\textbf{Short-term profile predominates:}} 
Jannach et al. \cite{jannach2015adaptation} have shown that the short-term intentions should be predominant in the selection of the recommendations. 
We can see in the first figure of Figure.\ref{fig:EDA}, the blue bar represents the mean percentage of user interactions hanging in the top 10 categories during the same session, and the orange one represents for that of items. This indicates that within one session, nearly half of interactions are in the main shopping target category, and 20 percent of clicks are for the target item. 

Overall both of them are subject to exponential decrease, which proves that user's short-term shopping goal plays a predominant role for the intra-session interaction choices.
Notice that Hidasi et al. \cite{hidasi2015session} propose GRU4REC, one early work on session-aware RNN-based recommender system, which avoids the cold start problem and significantly outperforms conventional baseline methods. 
From this point of view, we view RNN as one of the most advanced methods for short-term recommendation.
and choose it as the basis of our model design.

\item{\textbf{Long-term profile counts:}}  
Longer-term behavioral patterns and user preferences can also be important. In the middle figure of Figure.\ref{fig:EDA}, we plot the mean percentage value of a user clicking some repeated categories and items that he/she had clicked before in the previous 10 sessions. We can see that it tends to grow logarithmically and almost 60 percent of categories and 20 percent of items are repeated clicked inter previous 8 sessions. From this point of view, inter-session information contributes to 30 to 60 percent of information for next-basket category prediction and 5 to 20 percent of knowledge about repeated items. 

Several existing works have tried to use a simple static weighting strategy or hierarchical RNN \cite{jannach2015adaptation, jugovac2017efficient, quadrana2017personalizing, liu2018stamp, Meng2018WeaklySupervisedHT, tang2019towards} to combine the short-term and long-term models, but how to combine them in a seamless way still remains an open research problem. 
In order to enable long-term profiling, we propose an inter-session temporal dynamic model with anonymous session RNN model.

\item{\textbf{Time duration feature helps:}} 
One more common but not fully exploited feature is the click gap time, which is also the view dwell time of an item. This perfectly bridges the gap of discrete interaction sequence data with potential weights. Generally speaking, the longer time a user spends on the item, the more interest he has in it. According to the normalized histogram showed in Figure.\ref{fig:EDA}, the click gap of user interactions follows gamma distribution. Most users spend around 10 seconds between each click and normally the time duration is not longer than 5 minutes for each item. 

Such click gap time or item view duration helps us in connecting short-term and long-term profiles along time axis. So we design a novel triangle parallel attention network to incorporate temporal context in the RNN and perform efficient combination for short-term session sequence information. 

\end{enumerate}

Motivated by the above observations, in this paper, we want to quantify, exploit and integrate the effectiveness of user's intra-session and inter-session profiles with temporal dynamics.
First of all, since short-term profile plays a predominant role in user intent estimation, the very last actions in the present session should represent an important piece of context information to be taken into account when we make the recommendation.
Hidasi et al. \cite{hidasi2015session} propose GRU4REC, one early work on session-aware RNN-based recommender system, which takes these very last actions in users' intra-session sequence data. 
GRU4REC avoids the cold start problem and significantly outperforms conventional baseline methods. 
From this point of view, we view RNN as one of the most advanced methods for session-based short-term recommendation.
In order to maintain its short-term privilege, we choose session-based RNN recommender system as the basis of our model design.
However, as our exploratory data analysis shows above, long-term profiles are important for recommender system, while current state-of-art session-based approaches fail to model them effectively. 
Several existing works have tried to use a simple static weighting strategy or hierarchical RNN \cite{jannach2015adaptation, jugovac2017efficient, quadrana2017personalizing} to combine the short-term and long-term models, but how to combine them in a seamless way still remains an open research problem. 
In order to enable long-term profiling, we propose an inter-session temporal dynamic model with anonymous session RNN model. 
We choose to use an efficient embedding layer to automatically train and activate short and long term profiles from user embedding, short-term interest, user taste evolution and user survival time. For personalized recommendation, we add local negative sampling method: selecting negative samples in proportion to the item popularity within mini-batch sequences and ruling out the items appeared in his/her history.
Finally, we design a novel triangle parallel attention network to incorporate temporal context in the RNN and perform efficient combination for short-term session sequence information. 
Scott time binning method and extendable attention layer fully exert the role of temporal information.
In this way, user's item selection behavior can be predicted by mixed decision of short-term and long-term efficiently.

\begin{figure}[t] 
\includegraphics[width=0.45\textwidth]{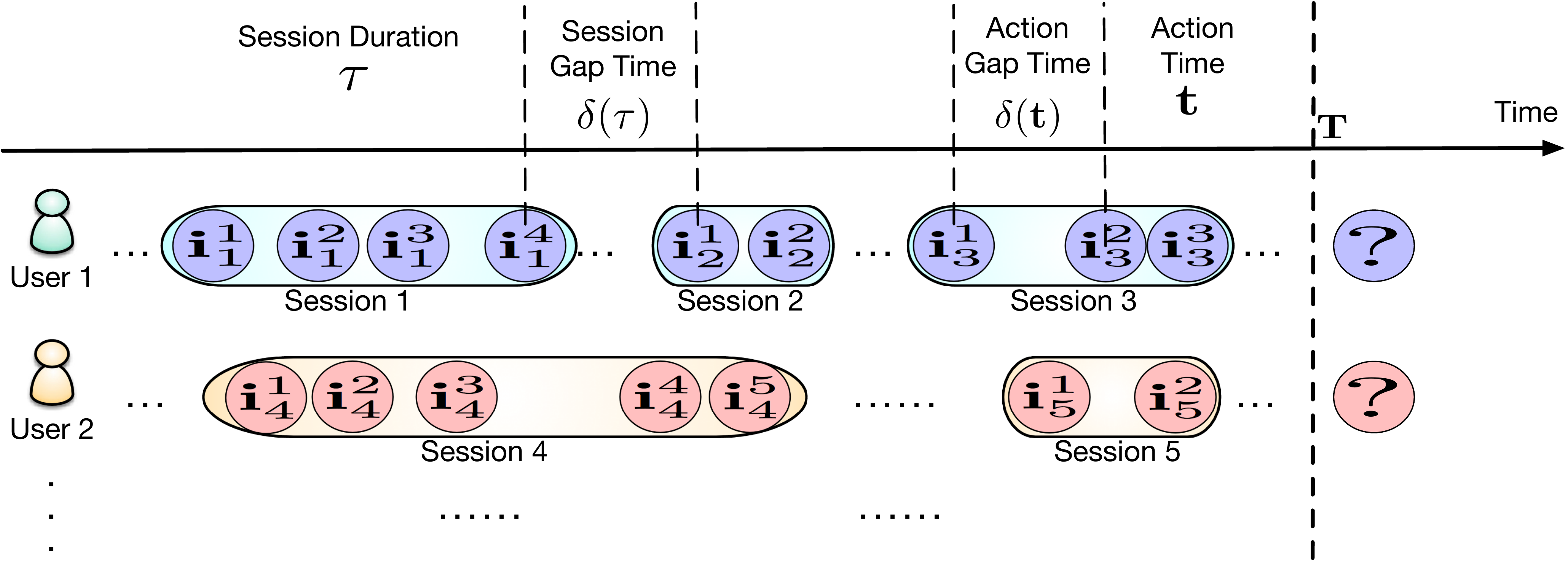}
\caption{Data flow of user and item interactions over time.}
\label{fig:dataflow}
\end{figure}

\subsection{Contributions}
The contributions of our ASARS framework can be described as follows:
\begin{enumerate}[leftmargin=*]

\item \textbf{Integrate Long-term by inter-session temporal dynamics model:} We include long-term user profiles for personalized session-based RS to learn the inter-session pattern by temporal dynamics model in a seamless way.
We integrate the time changes in session RNN and add user embedding, short-term interest, user taste evolution, user survival time and local negative sampling.

\item \textbf{Exploit short time by Triangle Parallel Attention Network:} We offer an novel attention model to exploit more intra-session information so as to enhance session-based RS in time dimension. We design a triangle parallel attention method for single sequence predicting and add a lower trigonometric transformation process to modulate the hidden states with multiplication efficiently. 

\item \textbf{Extensive Empirical Results:} We compete with four strong baseline models including BPR-MF (CF), YouTube (DNN), WaveNet (CNN) and GRU4REC (RNN) models and also compare five variants of our ASARS model. We conduct extensive experiments on four real datasets from different domains and demonstrate the effectiveness of ASARS for personalized recommendation. 
\end{enumerate}

\section{ASARS Framework}\label{model}

\begin{figure*}[t]
\centering
  \centering
  \includegraphics[width=.94\linewidth]{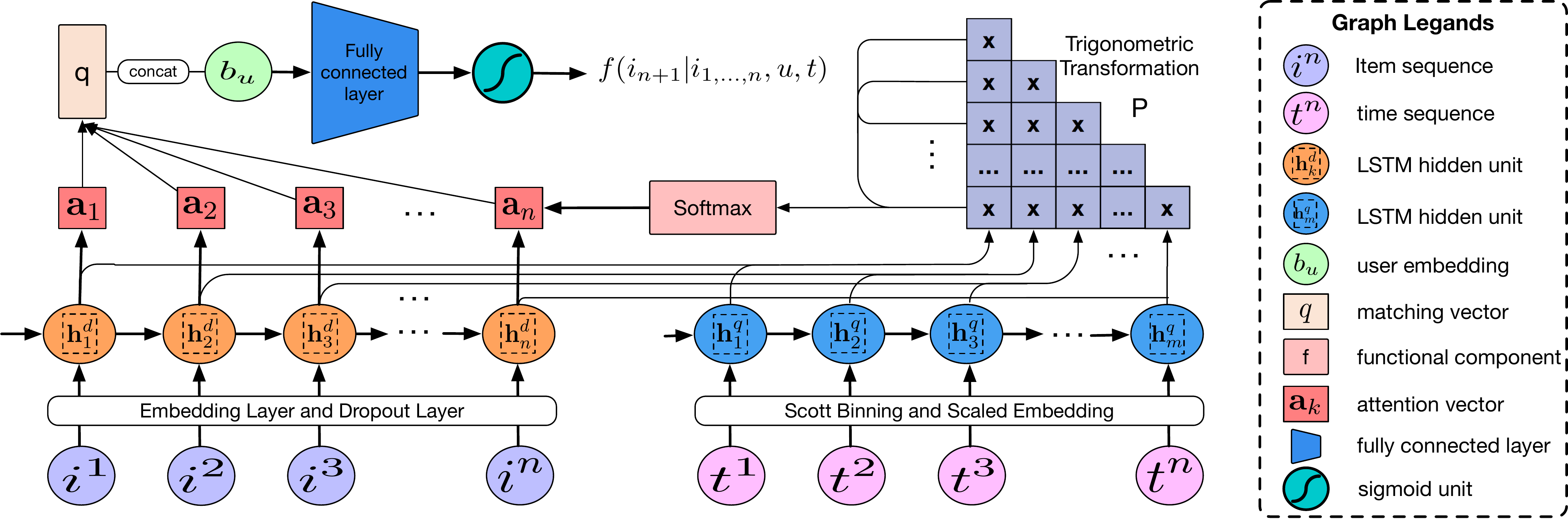}
  \captionof{figure}{Overall structure of ASARS model.}
  \label{fig:model}
 \end{figure*}

In this section, we describe our proposed personalized Attention Session-Aware Recommender System (ASARS) framework.
First, we introduce the session-based RNN framework in subsection \ref{sec:sessionRNN}.
Next, we explain how ASARS model combine short-time and long-term by inter-session temporal dynamics model in subsection \ref{sec:temporaldynamics}.
Then, ASARS model enhances the short-term profiles by using a a triangle parallel attention network layer \ref{sec:triangle} to sustain and exploit the inter-session patterns.
The overall structure of ASARS is shown as Figure.\ref{fig:model}

\subsection{Session-based RNN Framework} \label{sec:sessionRNN}
To show the concepts and notations clearly, we present an simplified data flow example in Figure.\ref{fig:dataflow}.
We define a user ``session'' as a set of continuous navigation activities without interruption in the log sequence.
In our settings, we separate each session by at least one-hour inactivity, which is commonly used in previous works \cite{wu2017session}.  
We denote a sequence of $m$ activity sessions as $S = \{ s_j | j=1, \dots, m\}$, where each session $s_j$ represents a user interaction event sequence $s_j = \{ i_j^1,   i_j^2,  \dots , i_j^{n_j} \}$.
Given a sequence of activity sessions, our goal is to predict what is the next item that the user mostly likely to interact with.
We formulate this as a ranking problem and solve it by first finding a scoring function $f(\cdot)$ that outputs the score of each item in the given the item list $I$, and then returning topK ranked items based on their scores. 
\begin{equation}
\hat{r}_k = f( i^{n}  |  i^{1, 2, \dots, k-1}), k \in I.
\end{equation}

First of all, in order to maintain the short-term predominant effect, our model is built on the session-based RNN model introduced in \cite{hidasi2015session}.
Session-based RNN model is based on LSTM/GRU layers and the hidden gates model the interaction order and relationship of user activities within a session. 

\begin{figure}[t]
\includegraphics[width=0.45\textwidth]{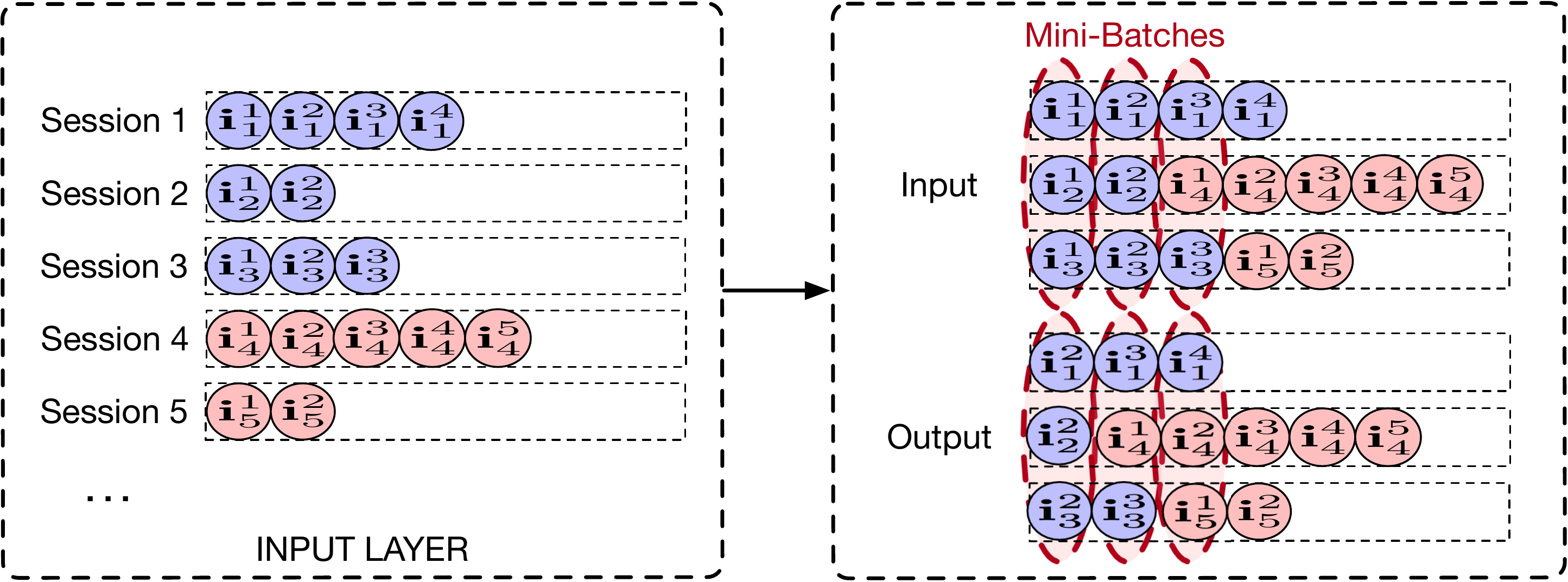}
\caption{Session-based RNN parallel mini-batch creation.}
\label{fig:minibatch}
\end{figure}

When processing a sequence, session-based RNN first input the sequence into the input layer, as shown at the left part of Figure.\ref{fig:minibatch}.

To deal with the various session length problem, it uses a session-parallel mini-batch approach to capture how a session evolves over the order of interacting activities \cite{hidasi2015session}.
If any of the sessions end, the next session is put behind of that sequence.
Note that this operation assumes all sessions are independent to each other.
Formally, we denote the $N_e$ mini-batched output sequences as
\begin{equation}
E = \{ e_{j} \} = \{ e_{j}^1, \dots,  e_{j}^{n_j}\} , j =1, \dots, N_e,
\end{equation}
where $e_{j, n}$ is the one-hot representation vector of the item.
Next, the one-hot mini-batch vector is fed into a GRU layer, and the hidden states are reset when switching sessions.
After that, the output of RNN can be treated as session-representations:
\begin{equation}
h_{session} = GRU(e_j, h_{session-1}), j =1, \dots, N_e.
\end{equation}
Finally, the last output of RNN gives the next step in this session, and the likelihood of being this item is calculated through a non-linear activation layer.
\begin{equation}
\hat{r}_k = g( e_k, h_{k} ) , k \in I.
\end{equation}
Normally, we use softmax, tanh or relu for the loss functions. 
There are some typical loss functions for recommender systems, like cross-entropy, BPR, and TOP1 loss proposed by the GRU4REC model.

Overall, session-based RNN is one of the state-of-art dynamic recommendation models which effectively exploits intra-session sequence order information.

However, sessions are not absolutely independent to each other, especially for task of personalized recommendation.
In the next subsection, we introduce how and why we design our model to effectively exploit inter-session patterns as well as temporal information. 

\subsection{Inter-Session Temporal Dynamics Model} \label{sec:temporaldynamics}

In traditional Matrix Factorization (MF) based approaches, the temporal dynamics model like \cite{koren2009collaborative} is commonly used for modeling time changes in data mining. Hereby, we model and learn the time changes by session RNN and user preferences as model drifting. 

Starting from the anatomy of a factor model with time changing factor:
\begin{equation} \label{eq:anatomy}
r_{ui} = e_i^T \cdot e_u + b_{u,i},
\end{equation}
where ${e_i}$ and $e_u$ are the one-hot representation vector of the items and user profiles, and $b_{u,i}$ represents the baseline predictor:
\begin{equation} \label{eq:predictor}
b_{u,i} = \mu + b_{u} + b_{i}.
\end{equation}
Here $b_{u}$, $b_{i}$ are the corresponding observed bias, and the overall average is denoted by $\mu$ .

\textbf{Time Changing.}
An illustrative data flow example is shown in Figure.\ref{fig:dataflow}.
We can see that there are four kinds of time information: action timestamp $t$, action gap time $\delta(t)$, session duration time $\tau$ and session gap time $\delta(\tau)$. 
Action timestamp can be used for periodical purchasing feature training directly as contextual information, and session gap time is helpful for survival analysis to predict user return time \cite{jing2017neural}.
Among all these temporal features, action gap time $\delta(t)$, also representing item dwell time, is the most valuable feature that haven't been fully exploited in previous session-based models. 
Therefore, adding time-changing factor to the Equation (\ref{eq:predictor}) and then it becomes:
\begin{equation} 
\hat{b_{ui}(t)} = \mu + b_{u}(t) + b_{i}(t).
\end{equation}

Then, we want to improve the session-aware recommender system by exploiting such item dwell time information.
Formally, for each session $j$, we create a dwell time sequence with the same dimension of item sequence as $t_j = \{ t_j^1,   t_j^2,  \dots , t_j^{n_j} \}$.
The item dwell time follows gamma distribution as shown in Figure.\ref{fig:EDA}. 
We can take bins of such time to reduce the dimensionality and then accelerate the training process.
We use Scott binning method \cite{scott1979optimal} for time feature such that the bin width is proportional to the standard deviation of the data and inversely proportional to cube root of original data size.
\begin{equation}
t_{bin} = \sigma \sqrt[3]{\frac{24*\sqrt{\pi}}{n}}.
\end{equation}
So the time model becomes 
\begin{equation} 
{b_{i}(t)} = b_{i}+ b_{i, t_{bin}}.
\end{equation}

Next, we use an embedding method to represent dwell time importance within sessions. 
\begin{equation}
E(t) = \{ e_{t,j} \} = \{ e_{t,j}^1, \dots,  e_{t,j}^{n_{t,j}}\},
\end{equation}
where $t = \{1, \dots, n_j$\}, $j = \{1, \dots, N_{j}\}$, $N_{j}$ is the number of users grouped by mini-batch size, and $e_{u,j}$ is the embedding vector of time. 
After training with a LSTM layer, instead of concatenating the hidden outputs directly, we explore to use attention scheme to integrate the timing effect to item sequence.
Intuitively, such attention vectors are perfectly used to modulate the outputs of hidden states representing session orders, and it's reported as a very useful tool to extract the importance of sequence vector. The triangle parallel attention network will be explained in section \ref{sec:triangle}.

\begin{figure}[t]
  \centering
  \includegraphics[width=.9\linewidth]{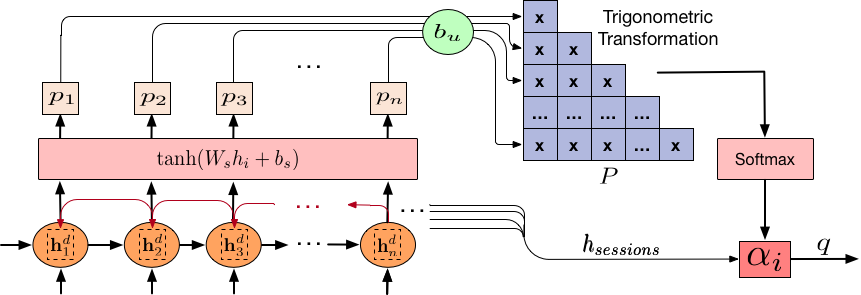}
  \captionof{figure}{Triangle parallel attention net with user profile.}
  \label{fig:attention}
\end{figure}

\textbf{Model Evolving.}
In addition, ASARS model also uses the long-term profiles by adding user embedding to learn the cross-session pattern and user favorite evolution as
\begin{equation} 
{e_{u}(t)} = e_{u}+ \alpha_u \cdot h_{u}(t),
\end{equation}
\begin{equation} 
{b_{u}(t)} = b_{u}+ \alpha_u \cdot dev_{u}(t),
\end{equation}
in which $e_{u}$ is user embedding, $b_u$ is the user bias, $h_{u}(t)$ shows the user short-term interest, $\alpha_u$ learns the user taste evolution and $dev_{u}(t)$ gives us the user survival time. 

Formally in ASARS, for user $u$, we denote the sessions grouped by users as $S(u) = \{ s_j | j=1, \dots, m_u\}$, where $s_j$ is the number $j_{th}$ session of user's total $m_u$ sessions.
Next, we use an embedding method to represent user's all behavioral patterns across sessions. 
Now we come to the sequence data preprocessing stage. 
We take the similar idea from parallel mini-batch method and change it to user-parallel mini-batch mechanism. 
Instead of complementing the dead end session row by the next session from all session lists, all mini-batches are selected and complemented within user's session groups.
So now we get the user-parallel mini-batched embedding sequence in input layer:
\begin{equation}
E(u) = \{ e_{u,j} \} = \{ e_{u,j}^1, \dots,  e_{u,j}^{n_{u,j}}\},
\end{equation}
where $u = \{1, \dots, m_u$\}, $j = \{1, \dots, N_{bu}\}$, $N_{bu}$ is the number of user grouped mini-batch size,  and $e_{u,j}$ is the embedding vector of item. 
Finally, we code the predictor as following:
\begin{equation}
\begin{split}
\hat{r_{ui}}(t) & = e_i^T \cdot (e_u+\alpha_u \cdot GRU \{ h_{session_{i-1}} \cdot dev_i(t)\} )  \\
 & + \mu + b_u + \alpha_u \cdot dev_{u}(t) + b_{i}+ b_{i, t_{bin}}.
\end{split}
\end{equation}

Such user representation aims to track user behavior patterns across sessions. 
There are many ways to combine new feature embeddings in neural network, such as concatenating features in embedding input layer directly, stacking two RNN layers for each feature respectively, co-training hierarchical RNN layers \cite{quadrana2017personalizing} and some more complex model structures like attention models \cite{yang2016hierarchical}, cross layers \cite{beutel2018latent}, memory networks \cite{sukhbaatar2015end} and meta-learning \cite{vartak2017meta}, etc.
Among them, we first tried to implement a simple model like concatenating or hierarchical RNNs.
Although the user embeddings in such simple model may not fully represent user behavior patterns,  these methods do make some improvements since more information have been included in training network and it can be trained faster.

\subsection{Triangle Parallel Attention Network}
\label{sec:triangle}

Notice that it's not straightforward to adapt sequence attention network directly to session-based RNN model. 
On one hand, traditional attention layer usually works with sequence-in-sequence-out RNNs in NLP tasks, but here we only predict one output sample in our recommender setting. 
On the other hand, in order to enable data augmentation \cite{tan2016improved} to get more training samples, all subsequences need to be forward to the attention network, such that the training time for forwarding process in attention network will be exponentially increased and make the model more difficult to train. 
Therefore, we design triangle parallel attention method for single sequence predicting and add a lower trigonometric transformation process to modulate the hidden states with multiplication efficiently.

As shown in the top right of Figure \ref{fig:model}, we introduce the time embedding $t^j$ in an attention network to reward items that play the most important role within session. 
The global attention mechanism yields the following formulas: 
\begin{equation}
\label{func:att1}
p_i = \tanh (W_s  h_{time} + b_s),
\end{equation}
\begin{equation}
\label{func:alpha1}
\alpha_i = \frac{e^{p_i^T h_{session}}}{\sum_{i} e^{p_i^T h_{session}}},
\end{equation}
\begin{equation}
\label{func:q1}
q_t = \sum_{i} \alpha_i h_{session},
\end{equation}
where $W_s$ and $b_s$ are parameters for training, and $h_{session}$ is the hidden output of item LSTM.

As mentioned above, the sequence weighting softmax and summation calculation cannot be adapted to the data augmentation training and will cause exponential training time cost. 
To accelerate this training process, we take lower trigonometric transformation to the vector $p_i^T b_u$ and forward it through the softmax function as a whole. 
In such a way, the training process can be hundreds of times faster.
Formally, for each hidden output $p_i, i = \{1, \dots, n\}$, we create an $n \times n$ lower trigonometric matrix $P$ with the sequence row $P_i$ as:
\begin{equation}
\label{func:tri}
P_i = [p_0^T \cdot b_u, \dots, p_i^T  \cdot b_u, 0, \dots, 0].
\end{equation}
After propagating such matrix $P$ through functions (\ref{func:alpha1}) and (\ref{func:q1}), we get the self-attention vector $\alpha$ and representation $q_t$.

Secondly, we also tried to use the attention network to combine the user embeddings with RNN outputs, as shown in Figure \ref{fig_attentionuser}. 
Similar to time attention scheme, we introduce user embedding $e_u$ in an attention network to reward session representations that are most favorite for the user. The self-attention mechanism yields the following formulas: 
\begin{equation}
\label{func:alpha}
\alpha_i = \frac{e^{p_i^T e_u}}{\sum_{i} e^{p_i^T e_u}}.
\end{equation}
After propagating through the attention layer or just embedding layer, we get the self-attention vector $\alpha$ and representation $b_u$.

Finally, the user attention vector weighted session representation $q$ concatenate with user representation $u$ and then goes to the following fully connected layer.
\begin{equation}
\begin{aligned}
\hat{r}_{j,k}  = g(q_t \cdot e_k  + b_j + b_k),
\end{aligned}
\end{equation}
where $g$ is a non-linear function for normalization, like softmax, tanh, relu, and etc.

\subsection{Improving Extensions}

\textbf{Loss functions:} We tried several common used loss functions in recommender systems, BPR \cite{rendle2009bpr}, TOP1\cite{hidasi2015session} and Hinge losses.
\begin{equation}
\text{BPR loss: } L = - \frac{1}{N_s} \sum_{j=1}^{N_s} \log{(\sigma (\hat{r}_j - \hat{r}_k))},
\end{equation}
\begin{equation}
\text{TOP1 loss: } L = \frac{1}{N_s} \sum_{j=1}^{N_s} \sigma (\hat{r}_j - \hat{r}_k) + \sigma(\hat{r}_j^2),
\end{equation}
\begin{equation}
\text{HINGE loss: } L = \max{ \{(\hat{r}_j - \hat{r}_k) + 1, 0 \}}.
\end{equation}

\textbf{Local negative sampling:}
Previous study has shown that negative sampling plays an important role in performance \cite{Shen2018MiningES}. 
Instead of random negative sampling, we also need to consider item popularity and user history issues. 
Specifically, we select negative samples in proportion to the item popularity within mini-batch sequences.
Furthermore, for each user, we need to rule out the items appeared in his/her history. 
This way, the local negative sampling method not only improves performance but also reduces the computational time as well.

\textbf{Data augmentation:}
Note that some users only have a few session histories, which may be insufficient for training the model with long-term user profiles.
So we need to make full use of all sequence samples and also their subsequences. 
First, we train each sequence with all hidden outputs and make the predictions, which fully explores the subsequences information.
Second, we leverage the dropout layer for the sequences such that it makes regularization as well as diversifies the input sequence data.

\section{Evaluation} \label{evaluation}

\begin{table}[t]
\small
  \caption{Dataset details.}
  \label{tab:dataset}
  \begin{tabular}{c|cccccc|ccl}
    \toprule
      Dataset & MovieLens & Recsys15 & Tianchi & OURS  \\
    \midrule
    Events &  53,309 & 17,920,066 & 6,921,446 & 254,398 \\  
    Users &  237 & / & 12,332 & 3,035\\  
    Items & 1,395  & 23,459 & 31,893 & 1,173\\  
   Sessions & 3,609 & 4,247,567  & 93,287 & 45,878\\
   \midrule
   Session support & 2 & 2 &  2 & 2\\
   Item support  & 10 & 20  & 10 & 20\\
   User support  & 10 & / &  10  & 20\\
    \bottomrule
  \end{tabular}
\end{table}
 
\begin{table*}[t]
\small
  \caption{Experimental Comparison Results -- shown are the MRR top 20 and Recall top 20 scores of four baseline models and five ASARS variants on four datasets. We highlight some focal improvements in bold and underline the best results. }
  \label{tab:dataset}
  \begin{tabular}{c|cc|cc|cc|ccl}
    \toprule
     \multirow{2}{*}{Models} & \multicolumn{2}{c}{MovieLens} & \multicolumn{2}{c}{Recsys15} & \multicolumn{2}{c}{Tianchi} &  \multicolumn{2}{c}{OURS}  \\
     & MRR@20 & RECALL@20 & MRR@20 & RECALL@20 & MRR@20 & RECALL@20 & MRR@20 & RECALL@20 \\
    \midrule
   BPR-MF CF &   0.004844 &  0.074627 & /  & /  &  0.001933 &  0.016234 &  0.015416  & 0.080431 \\  
   YouTube DNN &  0.014457 & 0.085271 & \textbf{0.194101}  &   0.499136 & 0.056148  &  0.139335   &  \textbf{0.025355}  & \textbf{0.103061} \\  
   WaveNet CNN & 0.010098   & 0.054264  &  0.100597 & 0.33733  &  \underline{\textbf{0.071221}} &  
\textbf{0.160209} &  0.023910   & 0.100067  \\  
   GRU4REC RNN &  \textbf{0.017358} &  \underline{\textbf{0.108527}} &  0.167908 & \textbf{0.570426}  &  0.049316 &  
0.127657  & 0.017474  & 0.058896 \\
   \midrule
   ASARS\_user\_att  & 0.012371 & 0.054264 &  /  &  /  &  0.041214  &  0.124636  & 0.015937  &  0.041002\\
   ASARS\_user\_cat & \textbf{0.018451}  & \textbf{0.100775} &  /  &  /  &  \textbf{0.053976} & 
\textbf{0.138174}  & \textbf{0.030365}  & 0.101227 \\
   ASARS\_time\_att  &  0.015988 & 0.038760 & \underline{\textbf{0.199309}} & \underline{\textbf{0.623005}}  &   
\textbf{0.057941} &  0.146510 &  \textbf{0.021176} & 0.067485 \\
   ASARS\_time\_cat &  0.017539 & 0.038760 &  0.181273  &  0.589828  &  0.054056 & 
0.140076  & 0.019368  & 0.061282 \\
   ASARS\_time\_user  & \underline{\textbf{0.020321}} & \textbf{0.100775} &  /  &  /  & 
\textbf{0.064585}  &  \underline{\textbf{0.204744}}  & \underline{\textbf{0.037259}}  & \underline{\textbf{0.106135}}  \\
    \bottomrule
  \end{tabular}
\end{table*}

\begin{figure*}[t] 
\includegraphics[width=0.23\textwidth]{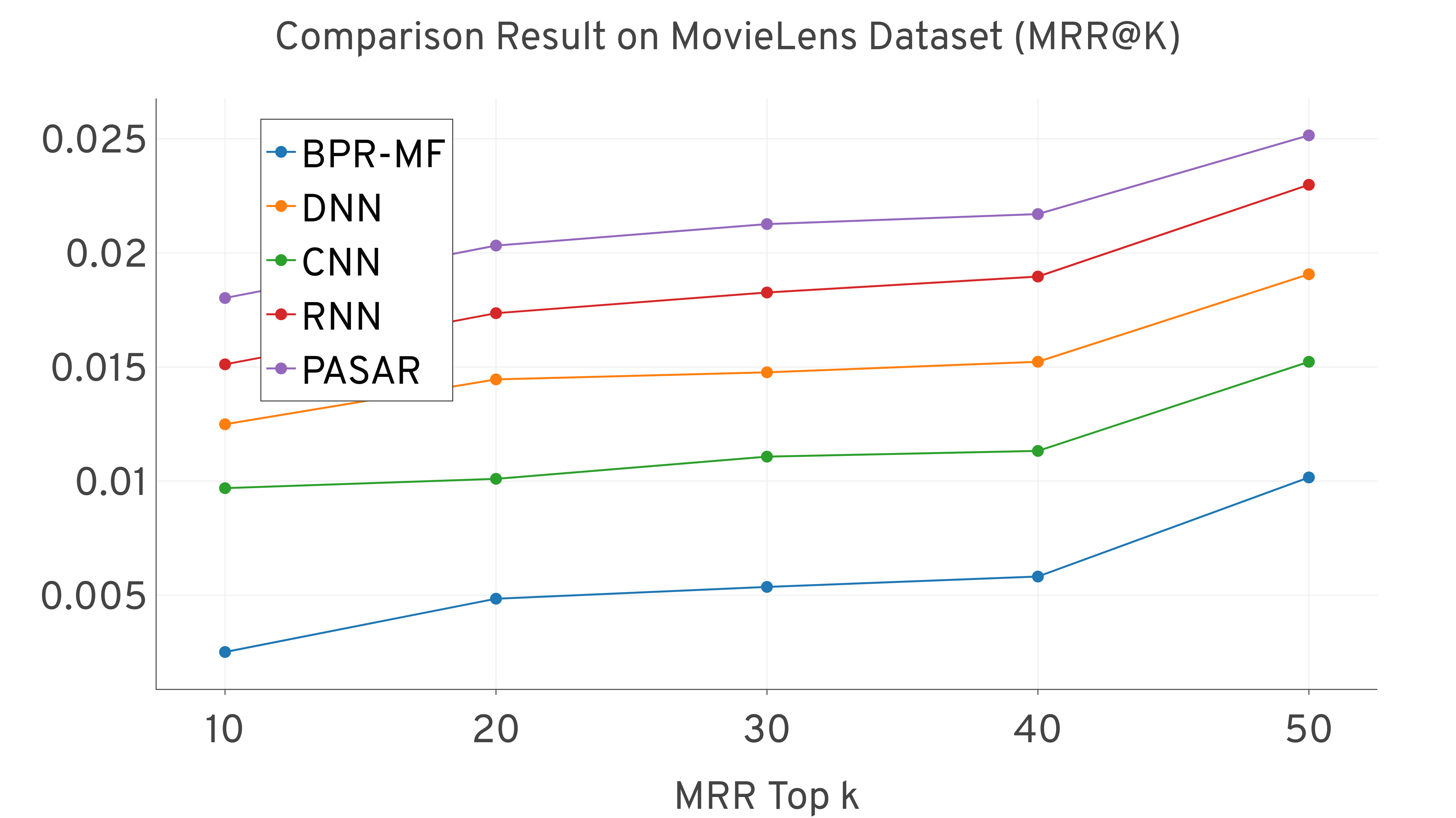}
\includegraphics[width=0.23\textwidth]{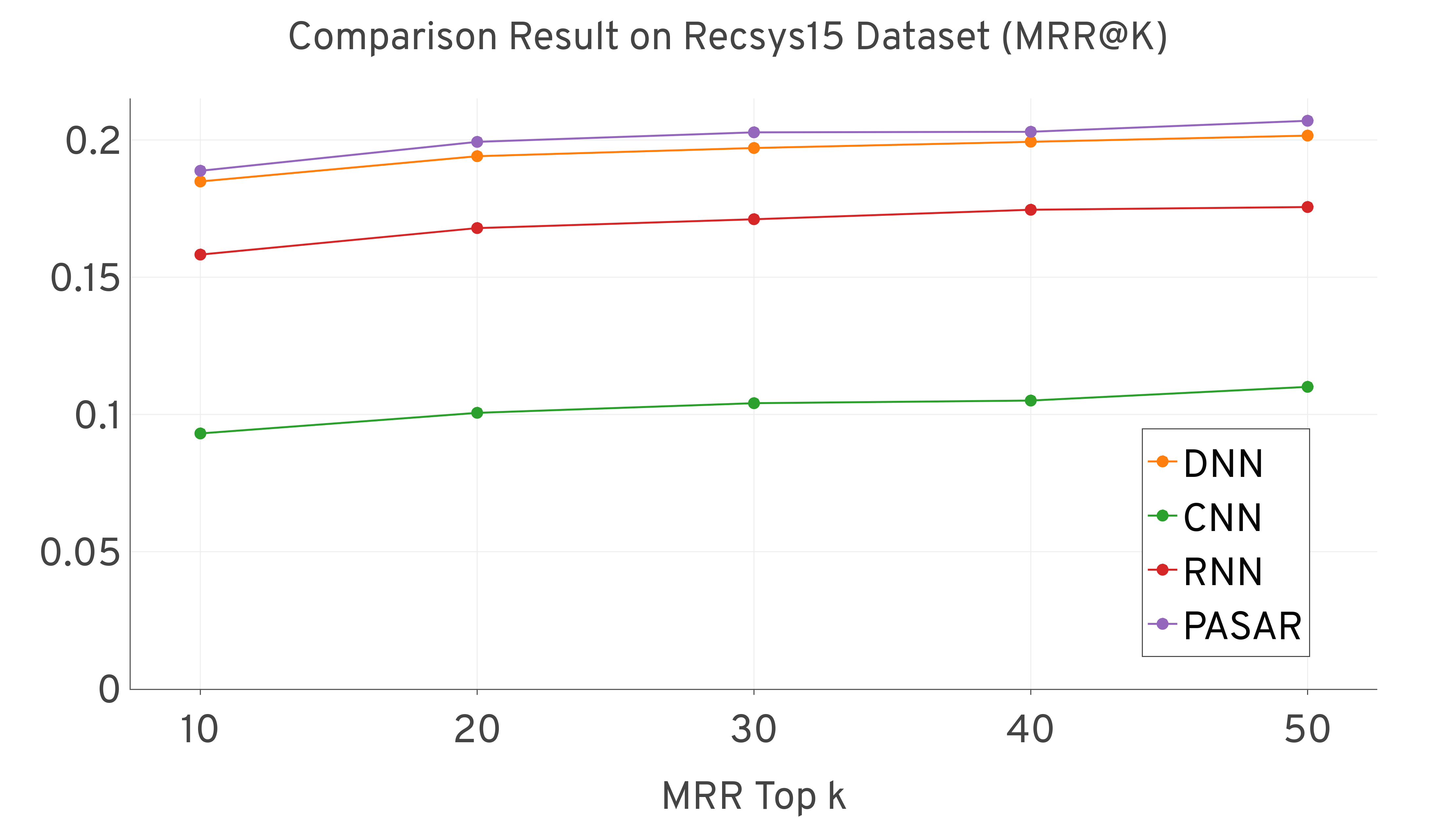}
\includegraphics[width=0.23\textwidth]{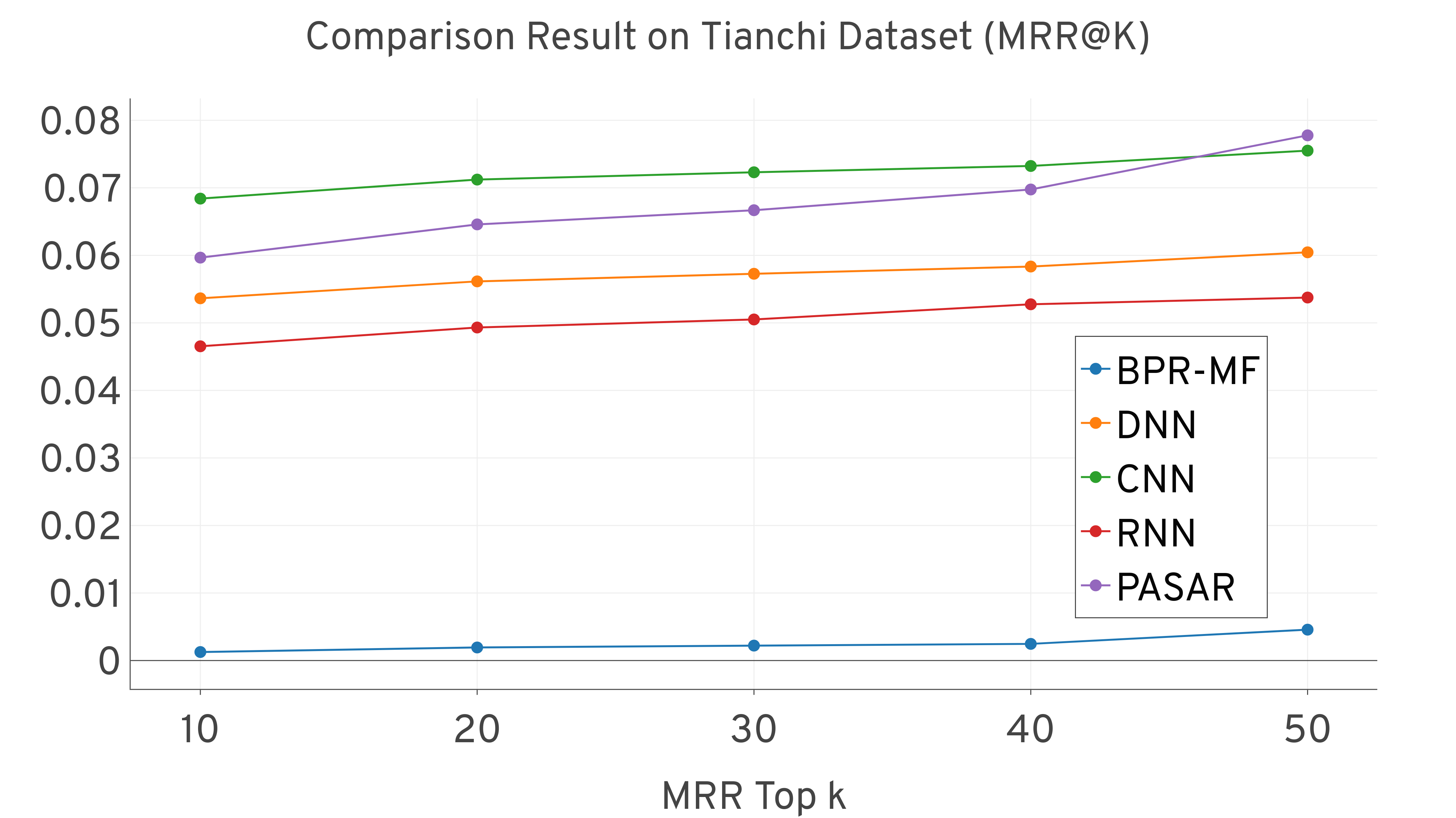}
\includegraphics[width=0.23\textwidth]{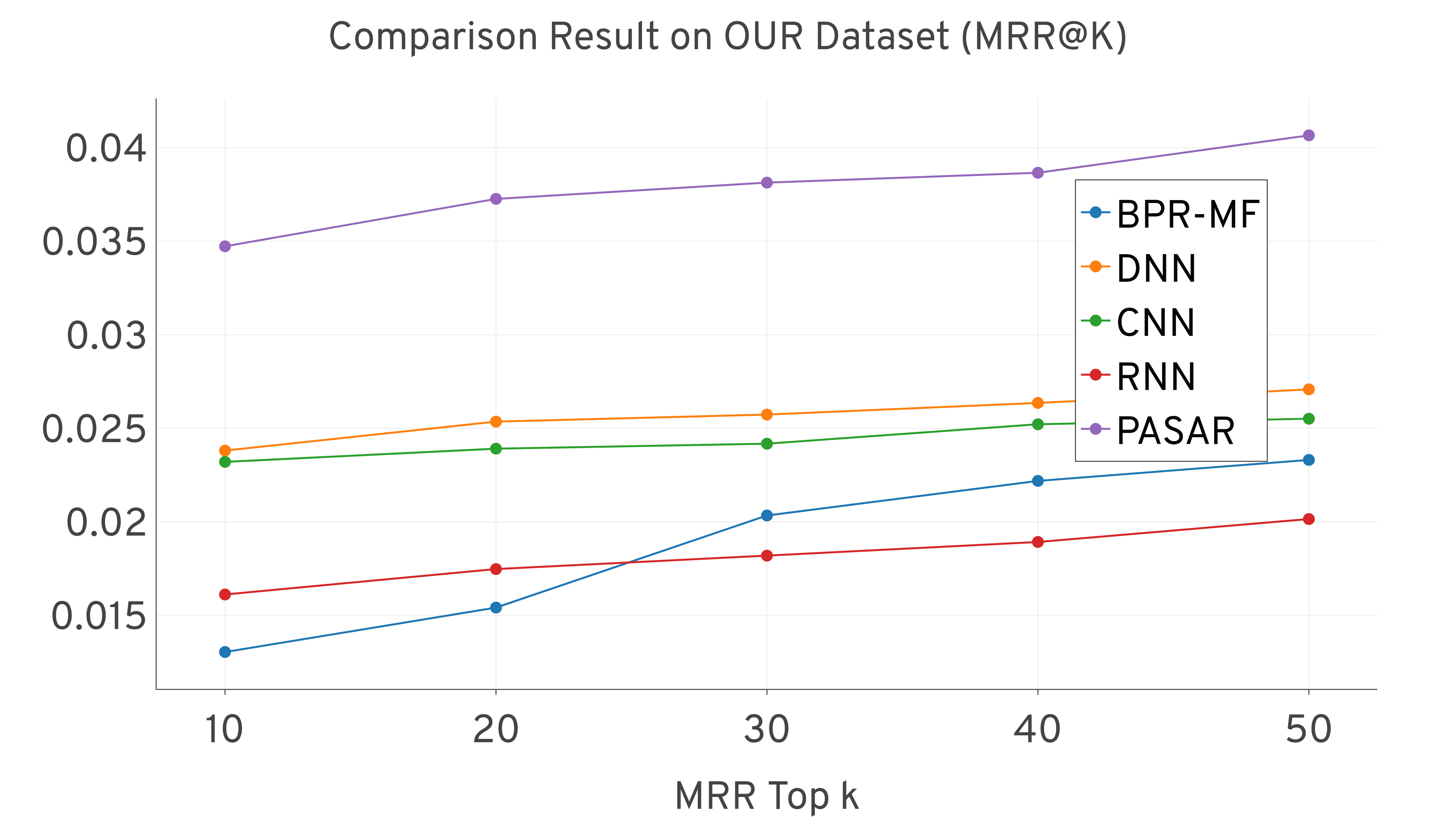}
\caption{Results of MRR@10, MRR@20, MRR30, MRR40 and MRR@all for MovieLens, Recsys15, Tianchi and our Datasets.}
\label{fig:mrr}
\end{figure*}

In this section, we will demonstrate the effectiveness and efficiency of our model for session-aware recommendation. 
First, we test and compare our model on multiple real-world datasets, from open source datasets to our own parsed real-world e-commerce dataset, covering both video and e-commerce domain. 
Second, we choose very strong baseline models to compete with, including traditional MF method and DNN, CNN, RNN based approaches. 
We describe our setup details and show the benchmarks and overall evaluation results as following.

\subsection{Datasets}

Totally we use four datasets in our experiments. 
The first is MovieLens \cite{harper2016movielens}, which is commonly used in recommender related works. 
The MovieLens 1M Dataset contains the ratings of 3,952 movies from 6,040 users from 2000 to 2003.
All users selected had rated at least 20 movies.
This movie rating data characterizes the user profiles in a extreme long term for 3 years. 
So we use this datasets mainly for concept proving of user long-term effects. 
The second is Recsys Challenge 2015 dataset \cite{ben2015recsys} that consists of 31,708,505 interaction events with 37,486 items in 7,981,581 sessions for 6 months.
This dataset only has the session info without user identities which session-based approaches commonly used, so we mainly test our time short-term effects on this dataset.
In order to test our model for both long and short term profiles, we choose Tianchi dataset \cite{yi2015purchase} containing 100M interaction events of 987,994 users with 4,162,024 items from 9,439 categories for 9 days. 
However, the duration is only 9 days which is not that long for user profiling. 
Finally, we also test on our newly clawed dataset from real-world e-commerce website containing 4,016,778 events for 126,468 users interacting with 390,381 items in 648,663 for two months.

Most of the datasets have no session ID info, we manually split the raw data into sessions based on 1-hour inactivity.
We add action gap time between each interaction timestamp within the same session and delete the last term. 
The most important preprocessing step is filtering the attributes with different support number. 
Since we add user long-term and time short-term to session-based model, we need to guarantee the user and item attributes have enough support training samples.
From our settings, we filter the items with at least 10 events, filter the sessions with length longer or equals to 2, and filter the users having 10 more sessions. 
To explore this supporting number influence, each dataset is split into sparse and dense two kinds of subsets for testing. 
All the dataset are partitioned to training and testing parts by cross validation based on both time and user. 
The test dataset contains sessions whose last timestamp is larger than a time boundary. 
We filter out the items and users that in the testing data but not in the training data.
The details of datasets are shown in Table.\ref{tab:dataset}.

\subsection{Comparing Baselines and ASARS Versions}

We compare ASARS with several strong baseline models. All of them are implicit ranking models.
First, we choose BPR-MF model \cite{koren2009matrix} representing CF-based approaches. 
Second, GRU4REC \cite{hidasi2015session} is the common baseline model for session-based RNN recommenders.
What's more, since deep neural network is very popular and powerful, we also compare with the YouTube recommender model \cite{covington2016deep} representing DNN approaches and WaveNet PixelRNN model \cite{van2016wavenet} representing Recurrent CNN models. 
Notably that most related session-based works didn't compare with DNN and CNN models previously and they only choose more CF-based methods as baselines. Especially for CNN models, from our knowledge there are seldom papers using recurrent CNN model for sequential recommendation task.
In some experimental settings, these methods are really competitive and show their advantages.
We will briefly introduce each baseline model and show the comparing results in the following sections.
\begin{itemize}
\item \textbf{BPR-MF} model \cite{koren2009matrix}: Matrix factorization techniques apply SVD factoring the user-item rating matrix, which are dominated in collaborative filtering recommenders. 
\item \textbf{YouTube DNN} model \cite{covington2016deep}: YouTube model includes two stages: candidate generation and ranking. 
\item \textbf{WaveNet CNN} model \cite{van2016wavenet}: PixelRNN aims to generate raw audio waveforms or phoneme recognition at first. Inner multiplicative relationships can be better exploited by its stacked causal atrous convolutions.
\item \textbf{GRU4REC RNN} model \cite{hidasi2015session}: We adapt GRU4REC model introduced in Section.\ref{sec:sessionRNN}. 
\end{itemize}

As for ASARS, we adapt user profile and dwell time in five different ways, two for user profile test, two for time feature test, and one for integrated version. The specifics of each model are as follows:

 \begin{itemize}
\item \textbf{ASARS\_user\_att} model: Adding user profile embedding by self-attention network. Based on session RNN, we add attention layer as Equation (\ref{func:alpha}) and propagate the mutual score by lower trigonometric transformation as Equation (\ref{func:tri}).
\item \textbf{ASARS\_user\_cat} model: Adding the user profile by directly concatenating the hidden outputs and the user embeddings, followed by a fully connected layer. 
\item \textbf{ASARS\_time\_att} model: Adding time profile embedding by global attention network as described in Section.\ref{sec:time}.
\item \textbf{ASARS\_time\_cat} model: Adding the user profile by directly concatenating the time gap embeddings and the item embeddings, and feeding into RNN layer as input sequences. 
\item \textbf{ASARS\_time\_user} model: Integrating both time and user profiles as final ASARS model as Figure.\ref{fig:model}. Comparing the design versions above, we choose to use attention net for dwelling time and concatenate user profiles.
\end{itemize}

\subsection{Implementation and Parameter Tuning}
We implement our model based on Spotlight \cite{kula2017spotlight}, an open PyTorch recommender framework. In this Spotlight model zoo, all IDs need to be regenerated mapping to continuous numerical IDs.
The model is trained end-to-end by back propagation.
In our model, we use single layer LSTM for item and time training. 
During the training process, we first grid search all the possible hyper-parameters optimized by Adagrad \cite{duchi2011adaptive} or Adam \cite{kingma2014adam}.
We also add early stop scheme when the evaluation loss does not decrease in the following 10 epochs. 
We evaluate the top-k ranking results using
MRR@K (Mean Reciprocal Rank) and Recall@K metrics. 
All metrics are the average of all  item lists in testing dataset.
The reciprocal rank is set to 0 if the rank is above K.

In our settings, all comparing baseline models and our model variants are trained by grid search and select the best result.
The best hyper-parameter set for ASARS\_time\_user model for MovieLens dataset is optimizing the hinge loss using Adagrad. The mini-batch size is 64. In the session information embedding, the maximum session sequence length is 200, the embedding size of item is 64, embedding size of time gap is 16 and embedding size of item is 32. The hidden dimension of LSTM layers are 100. The learning rate is set to 0.2. We used dropout regularization \cite{srivastava2014dropout} before the RNN layers with 0.5. For evaluation, we mainly focus on top 20 ranking results.

\subsection{Comparing Results}

All evaluation results are reported in Table.\ref{tab:dataset}. 
We mainly list the MRR top 20 and Recall top 20 scores of the four baseline models and five ASARS variants on the four datasets. We highlight some focal improvements in bold and underline the best results over all models. 
The detailed MRR@10, MRR@20, MRR30, MRR40 and MRR@all results for each datasets are shown in Figure.\ref{fig:mrr}
We finer analyze the comparing models by illustrating the user and time effects separately first, and then compare the overall performance. 

\textbf{Performance with User Long-term Profile:} 
Let's first study the long-term effects of user models, i.e. ASARS\_user\_att model and ASARS\_user\_cat model. 
Compared to the major baseline model GRU4REC, we can see that the concatenating method always outperforms the baseline for around 6\% improvement on Movielens, 9\% on Tianchi, and even 70\% improvement on our parsed dataset. 
This simple user model can exploit the long-term profile efficiently. 
However, our carefully designed user attention model does not perform well and some results even got worse to baseline model. 
Our motivation of designing such attention network for user profile is to learn the importance from the session sequence so that it can select the most influential items from previous item sequences for predicting. 
However, this scheme still cannot give better results after trying all kinds of model and hyperparameter tuning.
This may because users' favorites and behavior patterns vary a lot and hard to learn. 
What's more, although we guaranteed that all users have at least 10 session in training data, it still far from enough to train the attention network to work well. 
So user long-term profile is better to be used simply by concatenation model. 

\textbf{Performance with Time Short-term Profile:} 
We further investigate how the time short-term profile can be better exploited.
Comparing the results of ASARS\_time\_att model and ASARS\_time\_cat model, we can see that  ASARS\_time\_att works the best, which gives around 20\% improvement on Recsys15, Tianchi and our own datasets, except MovieLens data. 
This is expectable since MovieLens 1m data totally last for 3 years and the rating gap time cannot represent the info of the dwell time in e-commerce navigation sessions. 
We can see that with such useful addition info, ASARS\_time\_cat model can also beat the baseline model for about 10\% improvement.
Obviously, the global attention model for dwelling time helps more in session-based RNN model.

\textbf{Overall Performance with both User and Time:} 
The last ASARS\_time\_user model integrates the user concatenation model and time attention model, and it shows that with both long and short term info, our ASARS model can improve MRR@20 value about 30\% for Tianchi and 130\% for our dataset. 
Notably that DNN and CNN models performs better than our major baseline RNN model on Tianchi and our dataset. 
This shows that DNN and CNN models are more robust and general than RNN model for different recommender system settings. 
With the improving from our model design, ASARS can give the best performance and beat all other models.

\begin{figure}[t]
\centering
\begin{minipage}{.4\textwidth}
  \centering
  \includegraphics[width=.9\linewidth]{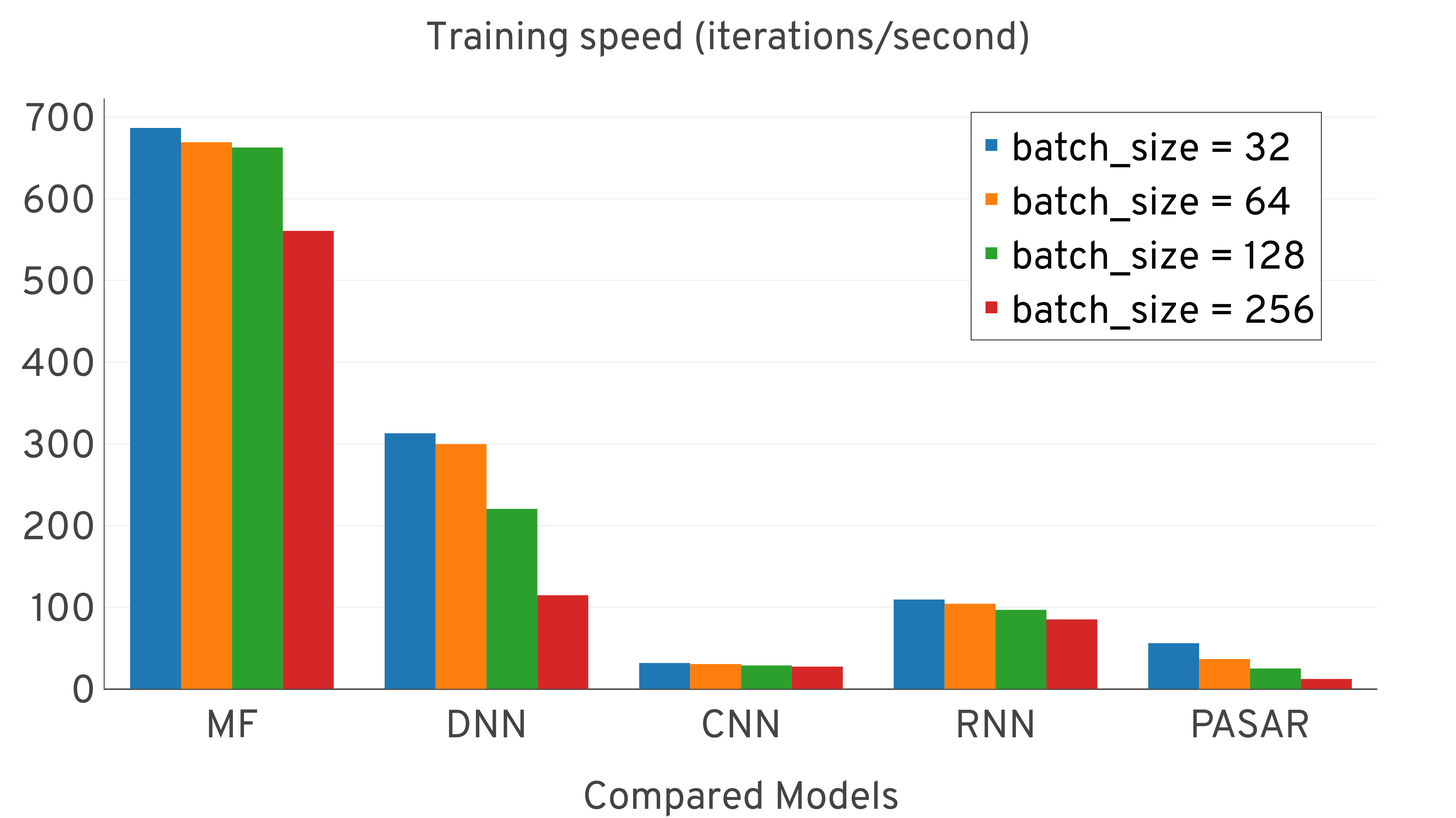}
  \captionof{figure}{Train speed (iter/s).}
  \label{fig:time}
\end{minipage}
\begin{minipage}{.4\textwidth}
  \centering
  \includegraphics[width=.9\linewidth]{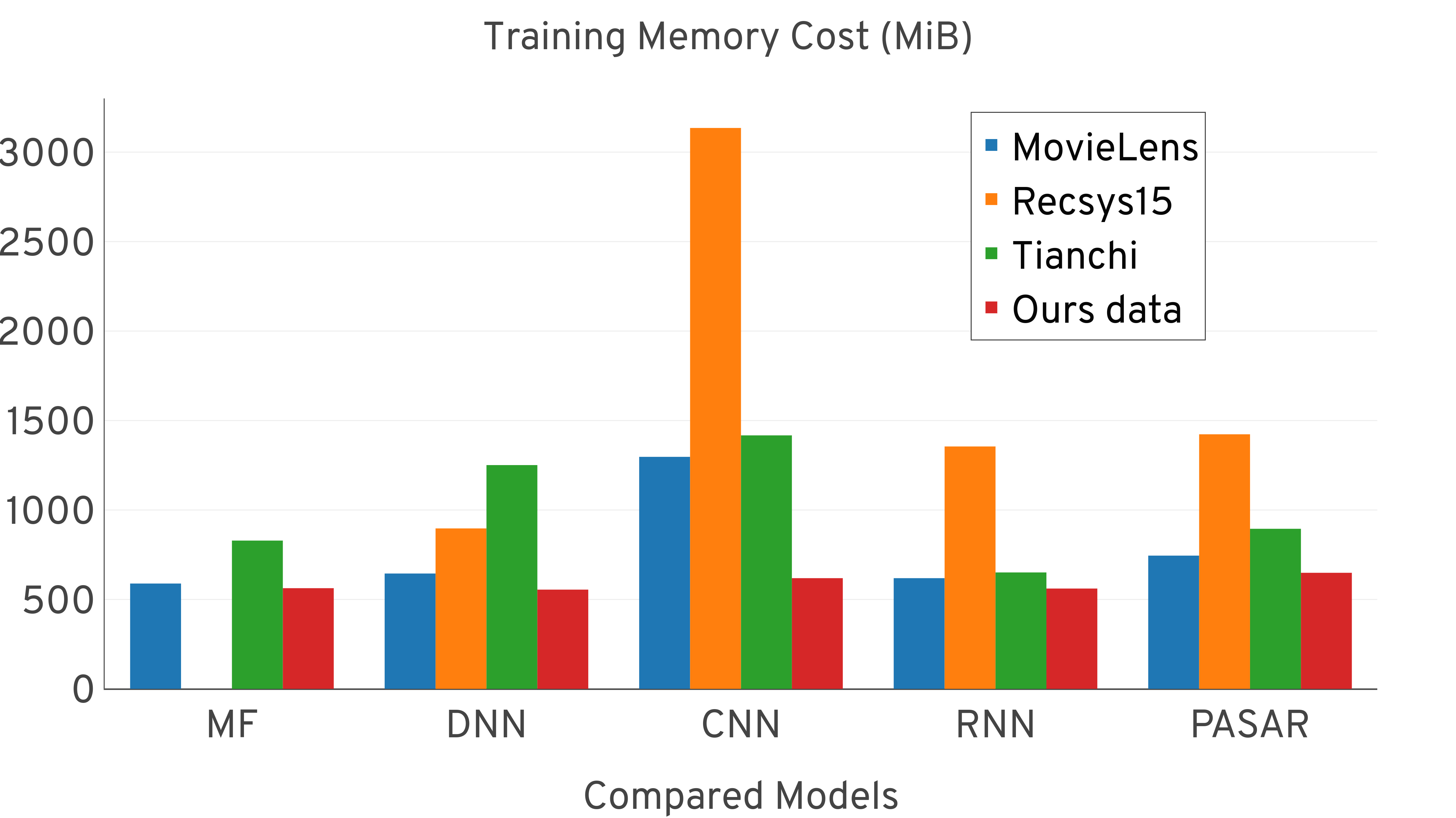}
  \captionof{figure}{Memory cost (MiB).}
  \label{fig:mem}
\end{minipage}
\end{figure}

\textbf{Memory and Time Cost:} 
Except for effectiveness, we also need to compare the memory and training time cost. 
We did experiments on NVIDIA Tesla P40 GPUs, and the training speed and memory cost are shown in Figure.\ref{fig:time} and Figure.\ref{fig:mem}. As expected, MF method is fastest and CNN method takes the most memory.
Our model is half slower than baseline RNN model and takes similar memory cost, which is acceptable for training process.

\section{Related Work} \label{relatedwork}

\begin{table*}[t]
  \caption{Related works compared by different methodology categories exploiting various domain features.}
  \label{tab:relatedworks}
  \begin{tabular}{cc|cccccccl}
    \toprule
     \multirow{3}{*}{Methods} & \multirow{3}{*}{Approaches} & \multicolumn{2}{c}{General} & Multiplicative & \multicolumn{2}{c}{Evolution} & Time & Sequence \\
     & \space & User Taste & Item Impression & Interaction & User Favorite & Item Trend & Drift & Info \\
      & \space & $p_u$ & $q_i$ & $b_{ui}$ & $b_u(t)$ & $q_i(t)$ & t & seq \\
    \midrule
    \multirow{3}{*}{CF-based} 
    & BPR-MF & \checkmark & \checkmark  & \checkmark & X & X & X & X \\  
    & TimeSVD++ & \checkmark & \checkmark & \checkmark & \checkmark & \checkmark & \checkmark  & X\\  
    & FPMC & \checkmark & \checkmark  & \checkmark & X & X & X & \checkmark \\  
    \midrule
    \multirow{3}{*}{NN-based} 
    & DNN & \checkmark & \checkmark  & \checkmark & X & X & X & X \\  
    & GRU4REC 
    	& X
 		& \checkmark & \checkmark  & X & X & X & \checkmark  \\  
    & ASARS & \textcolor{red}{\checkmark} & \textcolor{red}{\checkmark}  & \textcolor{red}{\checkmark} & \textcolor{red}{\checkmark} & \textcolor{red}{\checkmark} & \textcolor{red}{\checkmark} & \textcolor{red}{\checkmark} \\  
    \bottomrule
  \end{tabular}
\end{table*}

Much research efforts have been done to improve recommendation performance, like developing Context-Aware Recommendations \cite{adomavicius2011context}, Time-Aware Recommenders \cite{campos2014time} and Sequence-Aware Recommender Systems \cite{quadrana2018sequence}, exploiting contextual information, time dimension features and sequential order of the events. 
At the beginning, we list and compare some related research in different methodology categories exploiting various domain features in Table.\ref{tab:relatedworks}. 

\textbf{CF-based RS.}
Raised by the Netflix Prize \cite{bennett2007netflix}, factorization-based methods have been popularized and they framed the item-to-item recommender system, so called Collaborative Filtering method. 
Nowadays, kNN, SVD++ and BPR-MF \cite{rendle2009bpr, koren2010factor, rendle2012factorization} are still popular baseline methods for today's recommender research. 
With the motivation to profile temporal evolution of user and item favorites, TimeSVD++ \cite{koren2009collaborative} is one of the major works to add temporal dynamics to CF RS, modeling the factor model with time-changing feature $t$.
In addition to time, sequence prediction approach is further explored as well. FPMC \cite{rendle2010factorizing} is proposed to combine user-item matrix with Markov chains and it is still considered as one of the state-of-art sequential CF-based recommendation. 
Although CF-based methods have been theoretically well developed and are less expensive for computational cost, their practicalness and scalability yield to NN-based approaches.

\textbf{NN-based RS.}
As deep learning has been becoming in prominence in this decade, so as recommender system researchers began to apply deep neural network on recommendation.
One famous model is the YouTube DNN recommender \cite{covington2016deep}. 
It splits the recommendation task into two stages: a deep candidate generation model and a separate deep ranking model, and gives dramatic performance improvements.
Speaking of time or sequence modeling in NN, everyone would come up with Recurrent Neural Network, typically LSTMs and GRUs \cite{gers1999learning, chung2014empirical}.
Several recent works have been proposed to add temporal historical features for RNN recommender systems. 
Tims-LSTM \cite{zhu2017next} equips LSTM with time gates to model time intervals. 
RRN \cite{wu2017recurrent} endows both users and items with a LSTM autoregressive model that captures dynamics with a low-rank factorization.
NSR \cite{jing2017neural} uses survival analysis for return time prediction and exponential families for future activity analysis so as to solve the problem of Just-In-Time recommendation. 
Original and detailed survival analysis comes from temporal point process \cite{dai2016recurrent, du2015dirichlet, du2016recurrent}, which can recover both meaningful clusters and temporal dynamics.
Except for modifications based on LSTM, cross-layer scheme is another new-proposed way to discover contextual features more expressively \cite{wang2017deep, beutel2018latent}.
While these approaches did not adapt to session-based scheme, which could play a predominant actor for recommendation as shown in Section.\ref{sec:EDA}.

\textbf{Session-based RS.}
Session-based RNN recommender is first proposed by Hidasi et. al named GRU4REC \cite{hidasi2015session}. 
At first they mainly focus on anonymous cases and cold start problem in e-commerce recommender system, and they introduced session-parallel mini-batches RRN approach to fasten the training process. 
There are many follow-up papers based on that work: improving by data augmentation via sequence preprocessing \cite{tan2016improved}, exploiting dwell time by concatenating an additional dwell time RNN layer before item RNN \cite{dallmann2017improving}, adding different types of interactions and list-wise ranking framework \cite{wu2017session}, and personalizing it with hierarchical RNN \cite{quadrana2017personalizing}, etc. 
These works made incremental improvements for GRU4REC, but they do not make significant modification and haven't consider long-term intra-session info and user action gap time feature, which can make great gain according to Section.\ref{sec:EDA}
Most recently, the most related work is STAMP \cite{liu2018stamp}, but it has no use of dwelling time, no RNN, different attention scheme with not very impressing improvements.

\textbf{Long and Short-term Based.}
There have been plenty of works focusing on leveraging short-term features or long-term profiles in the past. 
Generally speaking, conventional Matrix Factorization based methods are more able to capture users' long-term general tastes \cite{hu2008collaborative}.
and it can be extended to detect their evolution trend with temporal dynamics \cite{koren2009collaborative}. 
STAR model \cite{song2015personalized} learned the long-term profile by Monte Carlo Markov Chain and used Latent Dirichlet Allocation for short-term profiling. Coupled tensor factorization proposed by \cite{rafailidis2015repeat} shows the repeat pattern in previous purchasing,
but RNNs show their privilege in short-term sequential pattern mining than other item-based or Markov Chain-based approaches.

Most recent RS for long and short-term sequential recommendation like \cite{devooght2017long, villatel2018recurrent}  also use RNNs, but they neglected the temporal info or not based on the session. It's impressing that Google just proposed a mixed model \cite{tang2019towards} almost integrate all my model variants, but my model is much lighter than that.
To facilitate RNN with long-term profiling, the goal of this paper is to make effective use of both long-term and short profiles and construct a better personalized session-aware RNN recommender system.

\section{Conclusion} \label{conclusion}

In this paper, we quantify, qualify and exploit the long-term user profile and short-term temporal dynamics for session-based RNN recommender systems.
In particular, we propose an Attentional Session-Aware Recommender System framework, called "ASARS", to integrate intra-session and inter-session profiles for both users and items with two novel models. 
We introduce inter-session temporal dynamic model to capture long-term user profiles for session-based RS to learn the inter-session pattern and user favorite evolution in a seamless way.
We design a triangle parallel attention network to leverage temporal dynamics scheme exploiting more intra-session time information so as to enhance session-based RS in time dimension.Such triangle parallel attention network is newly designed for sequence-in-single-out RNN structure and data augmentation needs, and also accelerate the training speed as well.
We demonstrate the improvement by our model design on four real-world datasets and beat comparable baseline models. 

\bibliographystyle{ACM-Reference-Format}
\bibliography{references}

\end{document}